\documentclass{article} 
\usepackage[preprint]{colm2026_conference}

\usepackage{microtype}
\usepackage{hyperref}
\usepackage{url}
\usepackage{booktabs}
\usepackage{subcaption}
\usepackage{graphicx}
\usepackage{algorithm}
\usepackage{algorithmicx}
\usepackage{algpseudocode}
\usepackage{amsmath}
\usepackage[capitalize,noabbrev,nameinlink]{cleveref}
\crefname{appendix}{Appendix}{Appendices}

\usepackage{lineno}

\usepackage{caption}
\usepackage{amssymb} 
\usepackage{xspace}

\definecolor{darkblue}{rgb}{0, 0, 0.5}
\hypersetup{colorlinks=true, citecolor=darkblue, linkcolor=darkblue, urlcolor=darkblue}
\newcommand{\AgentCE}{AgentCE-Bench\xspace}

\title{
\textcolor{red}{AgentCE}-Bench: \textcolor{red}{Agent} \textcolor{red}{C}onfigurable \textcolor{red}{E}valuation \\
with Scalable Horizons and Controllable Difficulty \\ under Lightweight Environments
}

\author{Wang Yang\textsuperscript{1}, Chaoda Song\textsuperscript{1}, Xinpeng Li\textsuperscript{1}, Debargha Ganguly\textsuperscript{1}, Chuang Ma\textsuperscript{2} \\\textbf{Shouren Wang\textsuperscript{1}, Zhihao Dou\textsuperscript{1}, Yuli Zhou\textsuperscript{3}, Vipin Chaudhary\textsuperscript{1}, Xiaotian Han\textsuperscript{1}} \\
\textsuperscript{1}Case Western Reserve University \textsuperscript{2}NII LLMC (Japan) \textsuperscript{3}University of Zurich\\
}

\begin{document}

\ifcolmsubmission
\linenumbers
\fi

\maketitle

\begin{abstract}
Existing Agent benchmarks suffer from two critical limitations: high environment interaction overhead (up to 41\% of total evaluation time) and imbalanced task horizon and difficulty distributions that make aggregate scores unreliable. To address these issues, we propose \AgentCE built around a unified grid-based planning task, where agents must fill hidden slots in a partially completed schedule subject to both local slot constraints and global constraints. Our benchmark offers fine-grained control through two orthogonal axes: \textbf{Scalable Horizons}, controlled by the number of hidden slots $H$, and \textbf{Controllable Difficulty}, governed by a decoy budget $B$ that determines the number of globally misleading decoy candidates. Crucially, all tool calls are resolved via static JSON files under a \textbf{Lightweight Environment} design, eliminating setup overhead and enabling fast, reproducible evaluation suitable for training-time validation. We first validate that $H$ and $B$ provide reliable control over task horizon and difficulty, and that \AgentCE exhibits strong domain consistency and model discriminability. We then conduct comprehensive experiments across 13 models of diverse sizes and families over 6 domains, revealing significant cross-model performance variation and confirming that \AgentCE provides interpretable and controllable evaluation of agent reasoning. Our code is available at
\url{https://github.com/uservan/AgentCE_Bench}.

\end{abstract}

\vspace{-10pt}
\begin{figure}[ht]
    \centering
    \begin{subfigure}{0.32\linewidth}
        \centering
        \includegraphics[width=\linewidth]{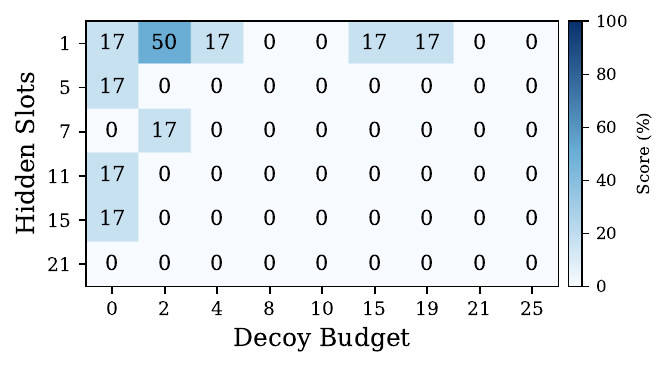}
        \caption{Qwen3.5-2B}
    \end{subfigure}
    \hfill
    \begin{subfigure}{0.32\linewidth}
        \centering
        \includegraphics[width=\linewidth]{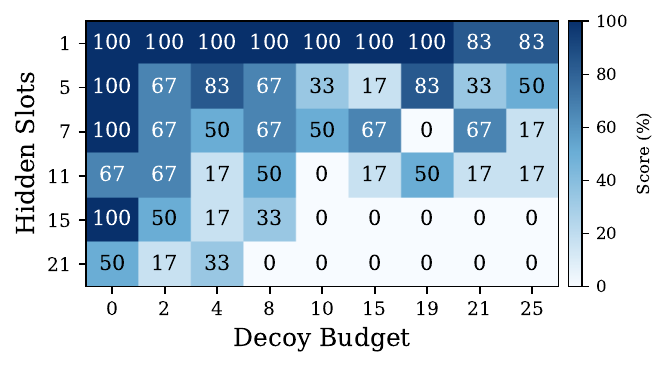}
        \caption{Qwen3.5-9B}
    \end{subfigure}
    \hfill
    \begin{subfigure}{0.32\linewidth}
        \centering
        \includegraphics[width=\linewidth]{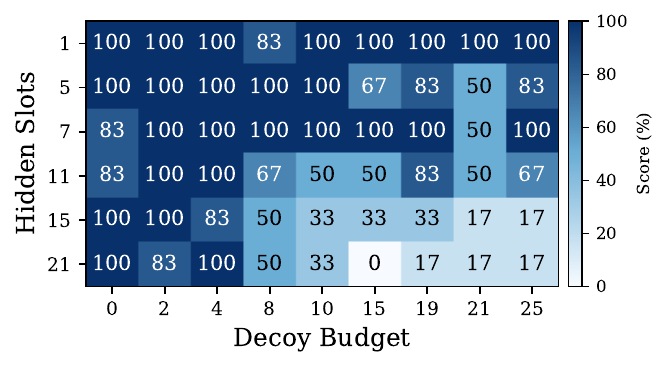}
        \caption{Qwen3.5-27B}
    \end{subfigure}
    \vspace{-8pt}
    \caption{
    This figure demonstrates the key benefits of \AgentCE as a benchmark: (1) it clearly distinguishes the agentic capabilities of models of different sizes, and (2) it identifies the capability boundary of each model under controlled agent task horizon and difficulty. Heatmaps of average reward (\%) for Qwen3.5 Dense models show that harder tasks (more hidden slots and larger decoy budget) yield lower rewards within each model, confirming configurable difficulty; larger models score higher, confirming discriminability across scales.
    }
    \label{fig:heatmaps_repr_qwen}
\end{figure}

\section{Introduction}

LLM-based agents have advanced rapidly in recent years, with models increasingly capable of executing complex, multi-step tasks across diverse real-world domains\citep{zhang2025landscape,schneider2025generative}. To keep pace with this progress, a rich ecosystem of agent benchmarks has emerged\citep{yehudai2025survey}---including WebArena\citep{zhou2023webarena}, TAU2-Bench\citep{barres2025tau,yao2024tau}, and TerminalBench~\citep{merrill2026terminal}---providing standardized environments to evaluate agent capabilities. Despite this progress, existing benchmarks suffer from two key limitations that hinder both evaluation efficiency and reliability.

First, many benchmarks adopt complex environment setups\citep{boisvert2024workarena++,deng2025swe,jimenez2023swe}---WebArena requires web simulators, TAU2-Bench relies on an LLM-as-User---causing environment interaction to consume 34\% and 41\% of total evaluation time, making them prohibitively expensive for large-scale or training-time evaluation. 

\begin{figure}[t]
\begin{center}
\includegraphics[width=\linewidth]{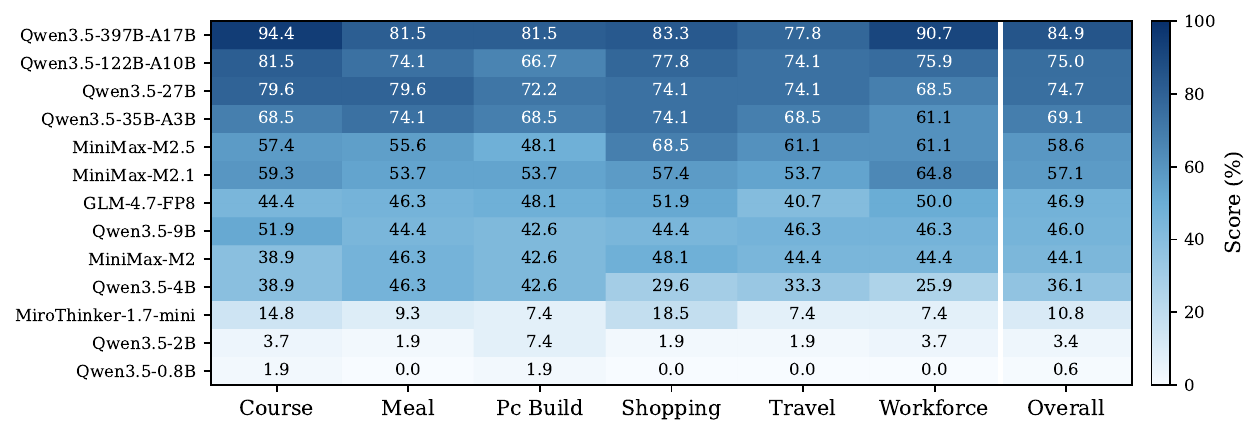}
\end{center}
\vspace{-12pt}
\caption{
Heatmap of average reward (\%) across models and domains. Each cell represents the average task completion reward of a model on a specific domain, highlighting cross-domain performance consistency and inter-model differences.
}
\label{fig:heatmap_models}
\end{figure}

\begin{figure}[t]
\begin{center}
 \includegraphics[width=\linewidth]{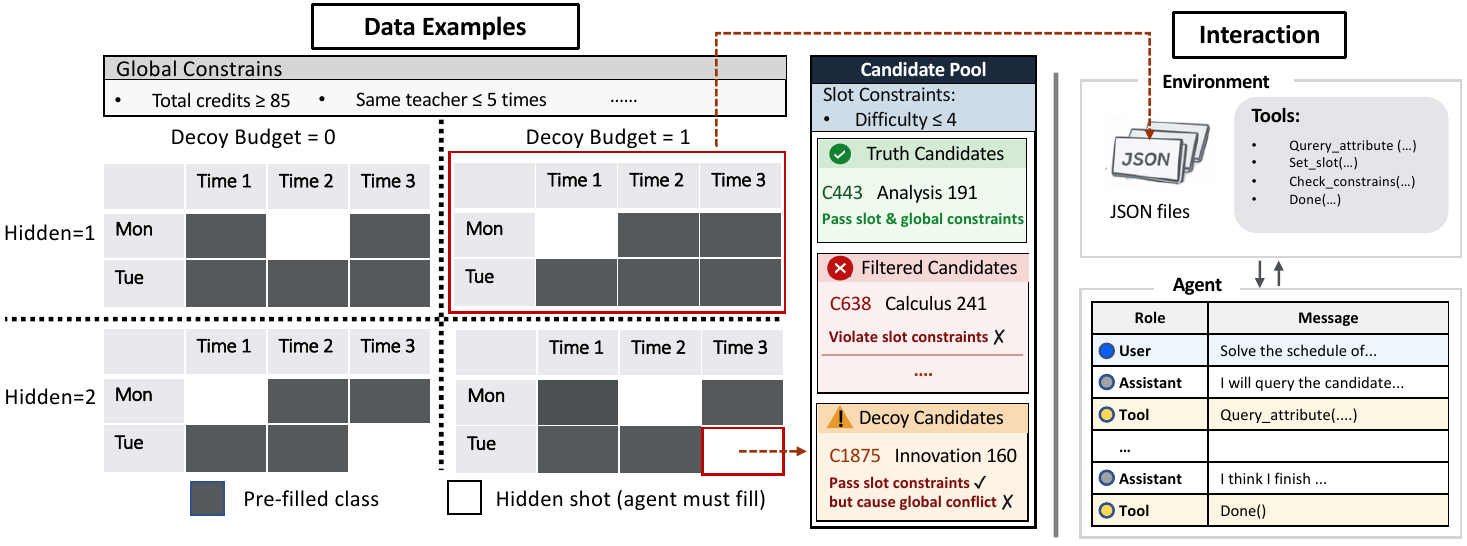}
\end{center}
\vspace{-5pt}
\caption{
Overview of \AgentCE framework. \textbf{Left (Data Examples):}  Each task is controlled by two configurable axes — hidden slots ($H$) and decoy budget ($B$). Each hidden slot has a candidate pool containing one truth (green), filtered candidates that violate slot constraints (red), and decoy candidates that pass slot constraints but cause global conflicts (orange). \textbf{Right (Interaction):} The agent operates in a lightweight JSON-based environment, iteratively querying and filling hidden slots through multi-turn tool-use dialogue, enabling precise measurement of agent capability across configurable difficulty levels.
}
\end{figure}

Second, existing benchmarks are highly imbalanced: task horizons vary widely across instances (e.g., from under 10 to over 100 interaction steps), and difficulty differs systematically across domains---reward scores range from $\sim$74--79\% on retail to $\sim$34--49\% on telecom---such that simple domain averaging masks true model capability. These limitations motivate a more principled evaluation framework that reduces environment overhead while accounting for uneven horizon and difficulty distributions.

To address these limitations, we propose \AgentCE built around a unified grid-based planning task, where agents must fill hidden slots in a partially completed schedule subject to both local slot constraints and global constraints across the entire grid. As illustrated in \Cref{fig:benchmark_illustration}, task instances are stored as static JSON files and agents interact through a lightweight tool interfAgentce, requiring no running services or external dependencies. 

Our benchmark offers three key advantages over existing benchmarks. \textbf{(1) Scalable Horizons}: the number of hidden slots $H$ directly controls task horizon, ranging from a single independent decision ($H{=}1$) to long-horizon reasoning chains ($H{=}17$), enabling systematic evaluation across varying steps. \textbf{(2) Controllable Difficulty}: the decoy budget $B$ independently governs difficulty by controlling the number of globally misleading decoy candidates, allowing fine-grained construction of instances from trivial ($B{=}0$) to highly challenging ($B{=}10$). \textbf{(3) Lightweight Environments}: unlike environment-heavy benchmarks, all tool calls are resolved by querying static files, eliminating setup overhead and making evaluation fast, reproducible, and suitable for training-time validation.

We first validate the effectiveness of \AgentCE through controlled experiments, confirming that hidden slots and decoy budget provide reliable and interpretable control over task horizon and difficulty, and that the benchmark exhibits strong domain consistency and model discriminability. Building on this, we conduct comprehensive experiments across 13 models of diverse sizes and families, as shown in \Cref{fig:heatmap_models}, revealing significant cross-domain and cross-model performance variation. Furthermore, \Cref{fig:heatmaps_repr_qwen} demonstrates that reward degrades consistently as hidden slots and decoy budget increase, confirming fine-grained controllability over both horizon and difficulty dimensions.

\section{Motivation: Limitations of Existing Agent Benchmarks}

\subsection{High Time and Resource Cost of Environment Interaction}

\begin{figure}[t]
\begin{minipage}{0.48\linewidth}
    \begin{center}
    \includegraphics[width=\linewidth]{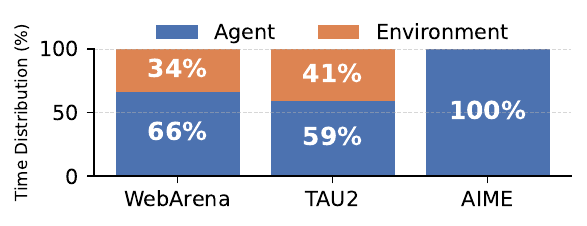}
    \end{center}
    \vspace{-13pt}
    \caption{Average time distribution of LLMs and environment interaction across three benchmarks (WebArena, TAU2, and AIME).}
    \label{fig:time_distribution}
\end{minipage}%
\hfill%
\begin{minipage}{0.48\linewidth}
    \begin{center}
    \begin{tabular}{ccc}
    \toprule
    \bf Benchmark & \bf Environment & \bf Resource \\
    \midrule
    tau2-bench & LLM-as-User & High \\
    WebArena   & Web Simulator & High \\
    AIME       & None & Low \\
    \bottomrule
    \end{tabular}
    \end{center}
    \captionof{table}{Environment type and resource requirements of three benchmarks (WebArena, TAU2-Bench, and AIME).}
    \label{tab:resource}
\end{minipage}
\end{figure}

Many existing benchmarks adopt complex environment setups to pursue realistic agent evaluation. 
For example, WebArena requires multiple web simulators, tau2-bench relies on an LLM-as-User to interact with the agent, and TerminalBench provides a collection of harbor-native tasks to evaluate terminal mastery. 
As shown in \Cref{tab:resource}, these environment-heavy benchmarks demand significant computational resources, whereas pure reasoning benchmarks like AIME require no external environment at all.


However, the pursuit of high environmental fidelity introduces substantial evaluation overhead. 
As illustrated in \Cref{fig:time_distribution}, we analyze the time breakdown between model inference and environment interaction across three benchmarks, using official logs from tau2-bench (GPT-4.1\citep{achiam2023gpt} as both agent and user) and WebArena (DeepSky Agent logs)\citep{zhou2023webarena}, alongside AIME\citep{balunovic_srimatharena_2025} as a reference.

\begin{figure}[t]
    \centering
    \begin{subfigure}[t]{0.32\linewidth}
        \centering
        \includegraphics[width=\linewidth]{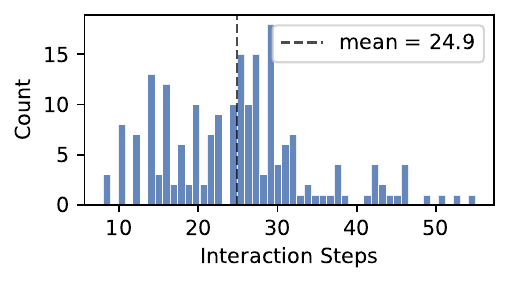}
        \caption{Airline}
        \label{fig:sub2}
    \end{subfigure}
    \hfill%
     \begin{subfigure}[t]{0.32\linewidth}
        \centering
        \includegraphics[width=\linewidth]{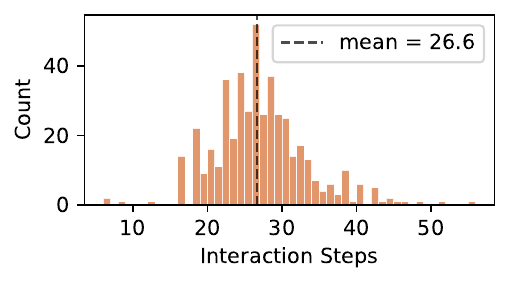}
        \caption{Retail}
        \label{fig:sub1}
    \end{subfigure}%
    \hfill%
    \begin{subfigure}[t]{0.32\linewidth}
        \centering
        \includegraphics[width=\linewidth]{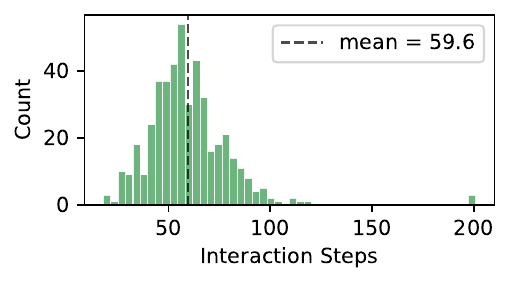}
        \caption{Telecom}
        \label{fig:sub3}
    \end{subfigure}%
    \vspace{0.5em}
    \begin{subfigure}[t]{0.32\linewidth}
        \centering
        \includegraphics[width=\linewidth]{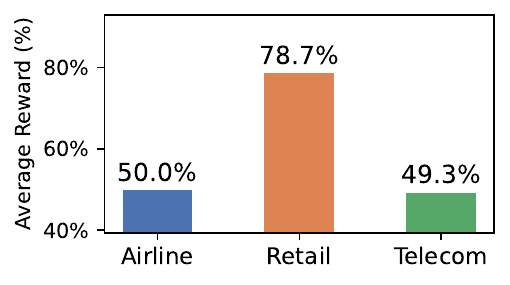}
        \caption{claude-3-7-sonnet}
        \label{fig:sub4}
    \end{subfigure}
    \begin{subfigure}[t]{0.32\linewidth}
        \centering
        \includegraphics[width=\linewidth]{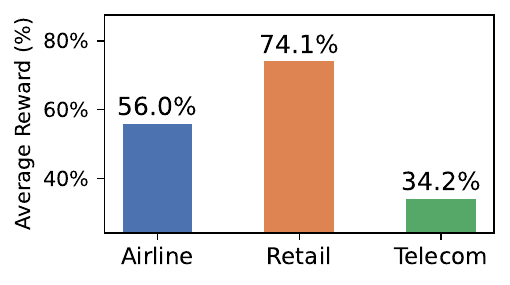}
        \caption{gpt-4.1}
        \label{fig:sub5}
    \end{subfigure}%
    \hfill%
    \begin{subfigure}[t]{0.32\linewidth}
        \centering
        \includegraphics[width=\linewidth]{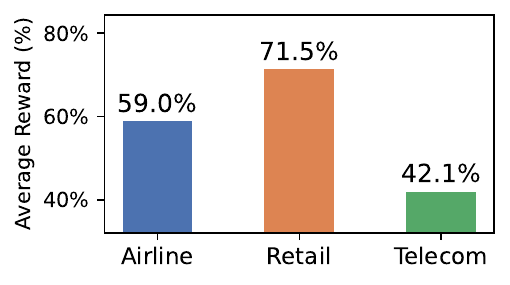}
        \caption{o4-mini}
        \label{fig:sub6}
    \end{subfigure}
    \vspace{-8pt}
    \caption{Top row: Distribution of task steps across three domains (airline, retail, and telecom), showing highly skewed and uneven distributions with large variance in task complexity. Bottom row: Reward scores by domain for Claude\citep{anthropic2024claude3}, GPT-4.1, and o4-mini, revealing significant difficulty imbalance across domains — some domains are consistently harder than others, with noticeably unequal performance distributions.}
    \label{fig:main}
\end{figure}


The results show that environment interaction accounts for approximately 34\% and 41\% of total evaluation time on WebArena and TAU2-Bench, while AIME spends nearly 100\% of its time on model inference since it requires no environment interaction.
This suggests that reducing environment interaction latency could significantly accelerate evaluation, and even enable the use of such benchmarks for validation during model training.

\subsection{Imbalanced Horizon and Difficulty Distribution}


As shown in \Cref{fig:main} (top row), the step distributions across all three domains are highly Imbalanced. Telecom tasks have a mean of 59.6 steps --- more than twice that of airline (24.9) and retail (26.6) --- with some exceeding 100 steps. Such imbalanced horizon distributions risk overweighting short tasks while underrepresenting long-horizon challenges.


Beyond horizon imbalance, task difficulty varies considerably across domains. As shown in \Cref{fig:main} (bottom row), retail is consistently the easiest domain ($\sim$74--79\% reward), while telecom is the hardest ($\sim$34--49\% reward) across all three models. This cross-domain difficulty gap means that a simple average over domains can be misleading --- a model may appear strong overall by excelling on easier domains while struggling on harder ones.

These two forms of imbalance expose a fundamental limitation of existing evaluation protocols: treating all tasks and domains equally can produce misleading conclusions about model capability. A model that solves many short, easy tasks may score well overall, yet fail systematically on the long-horizon, hard tasks that matter most in practice. This motivates the need for a more principled evaluation framework --- one that accounts for the uneven distribution of task horizons and adjusts for cross-domain difficulty, enabling a fairer and more diagnostic assessment of agent performance.

\section{Dataset Construction}

\subsection{Overview and Task Formulation.}

Each task instance consists of an $R \times C$ grid where each cell must
be filled with one item from a domain-specific pool $\mathcal{I}$.
A subset of $H$ cells are \textit{hidden slots} that the agent must
fill; the rest are pre-filled.
Each hidden slot is subject to \textit{slot constraints}
$\mathcal{C}_s$ (local attribute restrictions) and all slots must
jointly satisfy \textit{global constraints} $\mathcal{G}$ over the
entire grid.
Each hidden slot provides $K$ candidates, which fall into three
categories: the \textit{truth} (unique correct answer),
\textit{filter candidates} (violate $\mathcal{C}_s$, eliminable by
local reasoning), and \textit{decoy candidates} (satisfy
$\mathcal{C}_s$ but violate $\mathcal{G}$ under any valid
completion), ensuring a unique solution per instance.

\begin{figure}[t]
\begin{center}
\includegraphics[width=0.8\linewidth]{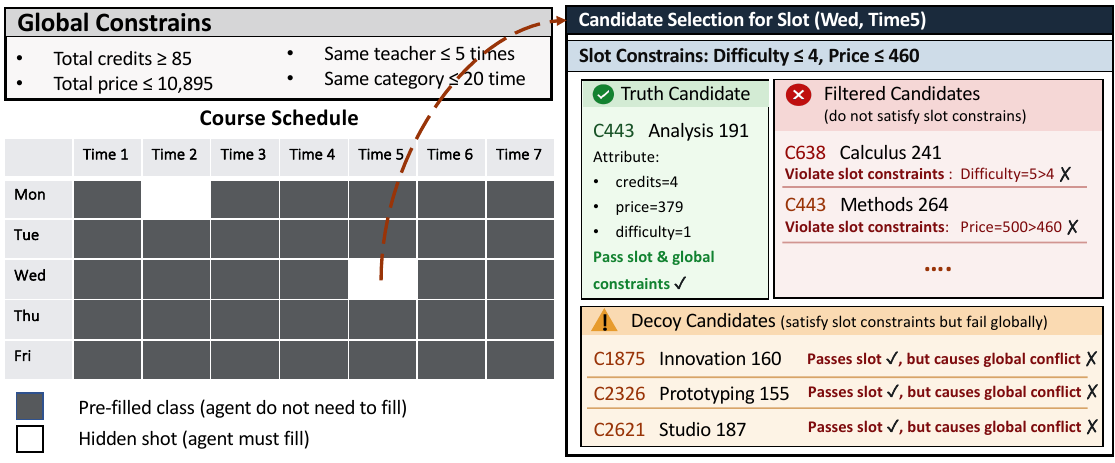}
\end{center}
\vspace{-8pt}
\caption{Illustration of a benchmark instance in the course scheduling domain. \textbf{Left:} A $5{\times}7$ course schedule grid with one hidden slot at (Wed, Time~5) that the agent must fill. Global constraints apply to the entire grid. \textbf{Right:} The candidate pool for the hidden slot, categorized into three types: \textit{Truth} (green) — the unique correct answer satisfying; \textit{Filtered Candidates} (red) — items that violate slot-level constraints; and \textit{Decoy Candidates} (orange) — items that pass slot constraints but cause global constraint violations}
\label{fig:benchmark_illustration}
\end{figure}

\noindent\textbf{Example shown in \Cref{fig:benchmark_illustration}.}
Consider a course scheduling instance with a $5 \times 7$ grid
($5$ days $\times$ $7$ time slots) and $H{=}5$ hidden slots.
The item pool consists of courses with attributes such as
\textit{credits}, \textit{price}, \textit{difficulty},
\textit{teacher}, \textit{category}, and \textit{workload}.
Global constraints include, for example:
total credits $\geq 85$, total price $\leq 10{,}895$,
and so on.
For hidden slot $(0, 1)$, the slot constraints require
difficulty $\leq 4$ and price $\leq 460$;
the truth answer is course \texttt{C443} (``Analysis 191'',
credits$=4$, price$=379$, difficulty$=1$).
Among the $24$ candidates, $17$ filter candidates
(e.g., \texttt{C723} with price$=433>460$ -- violating the local price limit)
are immediately eliminable by local constraints,
while $7$ decoy candidates (e.g., \texttt{C1146}, \texttt{C578})
satisfy the slot constraints locally but, when selected,
cause the global constraints to be violated regardless of
how the remaining slots are filled.

\subsection{Scalable Task Horizon via Hidden Slots}

The number of hidden slots $H$ directly controls the
\textit{task horizon}.
Each hidden slot requires the agent to query item attributes
and verify constraint compatibility before committing to a choice,
and since all slots share the same global constraints $\mathcal{G}$,
each decision constrains the feasible options for subsequent slots.
We vary $H \in \{1, 3, 5, 7, 9, 13, 17\}$: at $H{=}1$ the task
reduces to a single independent selection, while at larger $H$ the
agent must sustain coherent global reasoning across an increasingly
long decision sequence.

\subsection{Controllable Difficulty via Decoy Budget}

The \textit{decoy budget}
$B$ independently controls task difficulty by determining how many
globally misleading decoy candidates are introduced.
At $B{=}0$, no decoys are present and the task reduces to pure
local filtering; as $B$ increases, the agent must perform
increasingly extensive global constraint reasoning to distinguish
the truth from decoys.
We vary $B \in \{0, 2, 4, 6, 8, 10\}$ to construct instances
spanning a wide range of difficulty levels.

To generate decoys, we first select a subset
$\mathcal{S}_B \subseteq \mathcal{S}_H$ as \textit{decoy slots}
and partition $B$ into per-slot allocations $\{b_s\}_{s \in \mathcal{S}_B}$.
For instance, in the example above with $H{=}5$ and $B{=}15$,
four of the five hidden slots are selected as decoy slots with
allocations $[7, 5, 2, 1]$, meaning one hidden slot receives 7 decoys,
another hidden slot receives 5 decoys, and so on.

For each decoy slot $s$, we sample $b_s$ decoy candidates that
satisfy the local slot constraints $\mathcal{C}_s$ but are
guaranteed to violate $\mathcal{G}$ under any valid completion of
the remaining slots.
This guarantee is enforced during generation: a candidate is only
accepted as a decoy if, for all possible combinations of prior
decoy decisions, filling the remaining hidden slots with truth
answers still results in a global constraint violation.

As detailed in Appendix~\ref{appendix:data_generation}, targeted
sampling toward upper-bound constraints (e.g., total credits,
total price) is used to make this process tractable.

\subsection{Lightweight and Diverse Agent Scenarios}

To ensure broad coverage and avoid domain-specific biases,
our benchmark spans six real-world planning domains:
\textit{course scheduling}, \textit{grocery shopping},
\textit{travel itinerary}, \textit{workforce scheduling},
\textit{meal planning}, and \textit{PC build configuration}.
Each domain shares the same grid-based task structure, differing only in item attributes and
constraint semantics.

Unlike benchmarks that rely on interactive environments,
web browsers, or sandboxed executors, our benchmark is
\textit{lightweight}: all task data is stored in static JSON
files, and all tool calls are resolved by directly querying
these files without any running service or external dependency.
This design eliminates the overhead of environment setup and
makes evaluation fast, reproducible, and easily extensible
to new domains.

\subsection{Tool Settings}

Agents interact with each instance through a unified set of tools. For \textit{pre-filled slots}, the \textbf{item attribute query} tool returns full item attributes directly. For \textit{hidden slots}, only the \textbf{attribute filter query} tool is available, which returns candidate ids satisfying a specified condition (e.g., \texttt{price} $\leq 460$), preventing agents from retrieving all attributes at once and ensuring step-by-step reasoning under information constraints. The \textbf{slot constraint checker} and \textbf{global constraint checker} each return a boolean indicating whether local or global constraints are satisfied, with no diagnostic details. Finally, \textbf{set\_slot} fills or clears a hidden slot, and \textbf{done} signals task completion. The details are in Appendix~\ref{appendix:tool_system}.




\section{Experiments}

\subsection{Validation of the benchmark}

\begin{figure}[t]
    \centering
    \begin{subfigure}[t]{0.32\linewidth}
        \centering
        \includegraphics[width=\linewidth]{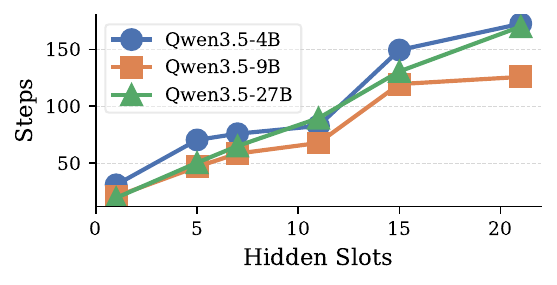}
        \caption{Horizon scaling: task steps vs.\ number of hidden slots.}
        \label{fig:horizon_scaling}
    \end{subfigure}
    \hfill
    \begin{subfigure}[t]{0.32\linewidth}
        \centering
        \includegraphics[width=\linewidth]{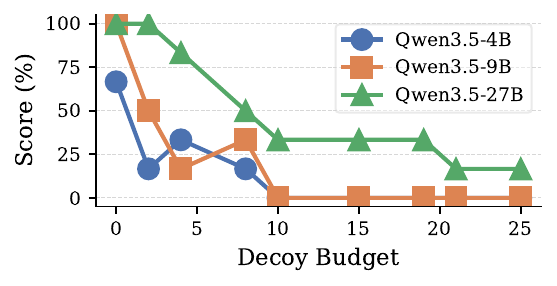}
        \caption{Difficulty control: score vs.\ decoy budget.}
        \label{fig:difficulty_control}
    \end{subfigure}
    \hfill
    \begin{subfigure}[t]{0.32\linewidth}
        \centering
        \includegraphics[width=\linewidth]{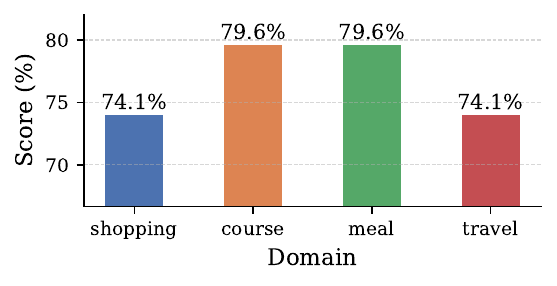}
        \caption{Domain-level average scores of Qwen3.5-27B.}
        \label{fig:discriminability_qwen}
    \end{subfigure}

    \vspace{0.5em}

    \begin{subfigure}[t]{0.32\linewidth}
        \centering
        \includegraphics[width=\linewidth]{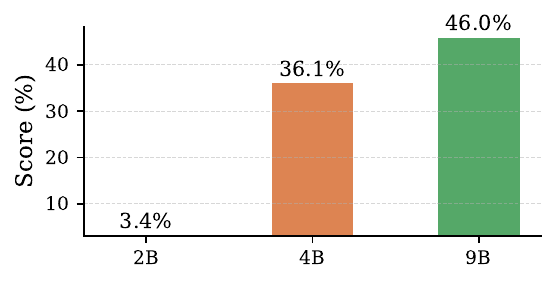}
        \caption{Model discriminability on Qwen3.5 Dense family.}
        \label{fig:discriminability_qwen_dense}
    \end{subfigure}
    \hfill
    \begin{subfigure}[t]{0.32\linewidth}
        \centering
        \includegraphics[width=\linewidth]{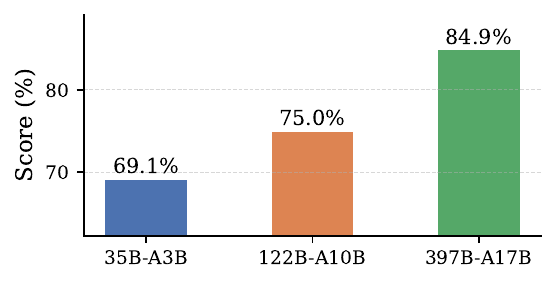}
        \caption{Model discriminability on Qwen3.5 MoE family.}
        \label{fig:discriminability_qwen_moe}
    \end{subfigure}
    \hfill
    \begin{subfigure}[t]{0.32\linewidth}
        \centering
        \includegraphics[width=\linewidth]{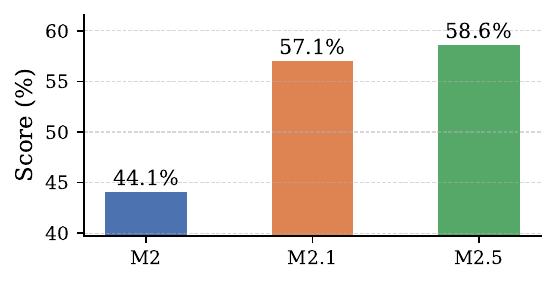}
        \caption{Model discriminability on MiniMax family.}
        \label{fig:discriminability_minimax}
    \end{subfigure}

    \vspace{-8pt}
    \caption{Verification of benchmark properties. (a) Task steps scale consistently with the number of hidden slots, confirming controllable horizon. (b) Score decreases as decoy budget increases, demonstrating effective difficulty control. (c) Domain-level scores of Qwen2.7B remain consistent across domains, suggesting low sensitivity to domain variation. (d--f) Score distributions across Qwen3.5 Dense, Qwen3.5 MoE, and MiniMax model families show strong discriminability, with rankings aligned with model scale.}
    \label{fig:verification_main}
\end{figure}
\textbf{Horizon Scaling}. To validate that the benchmark supports controllable task horizon, we fix the decoy budget to zero (i.e., no decoy candidates) and vary the number of hidden slots $h \in \{1, 5, 7, 11, 15, 21\}$. As shown in \Cref{fig:horizon_scaling}, the number of interaction steps increases consistently with $h$ across all three models (Qwen3.5-4B, 9B, and 27B)\citep{qwen3.5}, confirming that hidden slot count serves as an effective proxy for task horizon.

This is expected: for each hidden slot, the agent must query candidates by attribute conditions, reason over results, and call \texttt{set\_slot} to fill the slot. As $h$ increases, this process repeats more times, leading to a roughly linear growth in total steps, confirming that hidden slot count provides a principled and scalable control over task horizon. 

\textbf{Difficulty Control}. To validate difficulty controllability, we fix the number of hidden slots to $h{=}15$ and vary the decoy budget $b \in \{0, 2, 4, 8, 10, 15, 19, 21, 25\}$. As shown in \Cref{fig:difficulty_control}, task score decreases consistently as $b$ increases across all three models. 

This is because a larger decoy budget introduces more decoy candidates per slot --- items that satisfy local slot constraints but violate global constraints. As the number of such misleading combinations grows, the agent is more likely to select a decoy, causing global constraint violations and leading to lower scores. This confirms that decoy budget provides effective and continuous control over task difficulty.


\textbf{{Domain Consistency and Model Discriminability}}. We evaluate discriminability from two perspectives. First, in \Cref{fig:discriminability_qwen}, domain-level scores of Qwen3.5-27B remain consistent across domains (ranging from 74.1\% to 79.6\%), indicating that performance differences across domains are small and the benchmark does not introduce systematic domain bias.

Second, we examine whether the benchmark can distinguish models of different scales within the same family. As shown in \Cref{fig:discriminability_qwen_dense,fig:discriminability_qwen_moe,fig:discriminability_minimax}, scores increase monotonically with model size across all three families: Qwen3.5 Dense (3.4\% $\to$ 36.1\% $\to$ 46.0\%), Qwen3.5 MoE (69.1\% $\to$ 75.0\% $\to$ 84.9\%), and MiniMax\citep{chen2025minimax} (44.1\% $\to$ 57.1\% $\to$ 58.6\%). This consistent scaling behavior demonstrates that the benchmark provides strong discriminability and reliably reflects model capability across architectures and scales.


\subsection{Main Results across Diverse LLMs}

\begin{table}[t]
    \centering
    \resizebox{\linewidth}{!}{%
    \begin{tabular}{cc|ccc|ccccccc}
        \toprule
        \multicolumn{2}{c|}{} & \multicolumn{3}{c|}{\textbf{Usage} {\scriptsize(steps / Hour / K)}} & \multicolumn{7}{c}{\textbf{Score (\%)}} \\
        \cmidrule(lr){3-5} \cmidrule(lr){6-12}
        \textbf{Arch.} & \textbf{Size} & \textbf{Steps} & \textbf{Time} & \textbf{Tokens}
        & \textbf{Course} & \textbf{Meal} & \textbf{PC} & \textbf{Shop} & \textbf{Travel} & \textbf{Work} & \textbf{Avg.} \\
        \midrule
        \multicolumn{12}{c}{\textit{Qwen3.5 Dense}} \\
        \midrule
        Dense & 0.8B & 283.5 & 3.9 & 67.1 & 1.9  & 0.0  & 1.9  & 0.0  & 0.0  & 0.0  & 0.6  \\
        Dense & 2B   & 174.4 & 3.4 & 61.9 & 3.7  & 1.9  & 7.4  & 1.9  & 1.9  & 3.7  & 3.4  \\
        Dense & 4B   & 144.5 & 2.4 & 29.8 & 38.9 & 46.3 & 42.6 & 29.6 & 33.3 & 25.9 & 36.1 \\
        Dense & 9B   & 134.1 & 2.0 & 24.2 & 51.9 & 44.4 & 42.6 & 44.4 & 46.3 & 46.3 & 46.0 \\
        Dense & 27B  & 175.4 & 5.2 & 19.1 & 79.6 & 79.6 & 72.2 & 74.1 & 74.1 & 68.5 & 74.7 \\
        \midrule
        \multicolumn{12}{c}{\textit{Qwen3.5 MoE}} \\
        \midrule
        MoE & 35B-A3B   & 164.6 & 4.4  & 31.4 & 68.5 & 74.1 & 68.5 & 74.1 & 68.5 & 61.1 & 69.1 \\
        MoE & 122B-A10B & 156.0 & 3.8  & 21.8 & 81.5 & 74.1 & 66.7 & 77.8 & 74.1 & 75.9 & 75.0 \\
        MoE & 397B-A17B & 233.5 & 10.2 & 25.2 & 94.4 & 81.5 & 81.5 & 83.3 & 77.8 & 90.7 & 84.9 \\
        \midrule
        \multicolumn{12}{c}{\textit{MiniMax}} \\
        \midrule
        MoE & M2-229B   & 357.0 & 5.1  & 23.8 & 38.9 & 46.3 & 42.6 & 48.2 & 44.4 & 44.4 & 44.1 \\
        MoE & M2.1-229B & 453.7 & 5.2  & 35.7 & 59.3 & 53.7 & 53.7 & 57.4 & 53.7 & 64.8 & 57.1 \\
        MoE & M2.5-229B & 436.1 & 5.8  & 32.4 & 57.4 & 55.6 & 48.2 & 68.5 & 61.1 & 61.1 & 58.6 \\
        \midrule
        \multicolumn{12}{c}{\textit{MiroThinker}} \\
        \midrule
        Dense & mini-30B & 227.5 & 28.0 & 97.9 & 14.8 & 7.4  & 7.4  & 18.5 & 3.7  & 9.3  & 10.2 \\
        \midrule
        \multicolumn{12}{c}{\textit{GLM-4.7-FP8}} \\
        \midrule
       MoE & 358B & 190.9 & 5.5 & 21.4 & 44.4 & 46.3 & 48.2 & 51.9 & 40.7 & 50.0 & 46.9 \\
        \bottomrule
    \end{tabular}%
    }
    \vspace{-8pt}
    \caption{Evaluation results of different models across six domains. \textbf{Arch.} denotes the model architecture (Dense or MoE). \textbf{Steps}, \textbf{Time} (total hours on 8$\times$H200), and \textbf{Tokens} (K) report the average interaction steps, total wall-clock time, and average completion tokens per task, respectively. The benchmark consists of $6 \times 9$ tasks per domain, with hidden slots sampled from $\{1, 5, 7, 11, 15, 21\}$ and decoy budgets from $\{0, 2, 4, 8, 10, 15, 19, 21, 25\}$. Domain scores and \textbf{Avg.} report the average reward (\%) per domain and overall.}
\label{tab:main_results}
\end{table}

\begin{figure}[t]
    \centering
    \begin{subfigure}{0.32\linewidth}
        \centering
        \includegraphics[width=\linewidth]{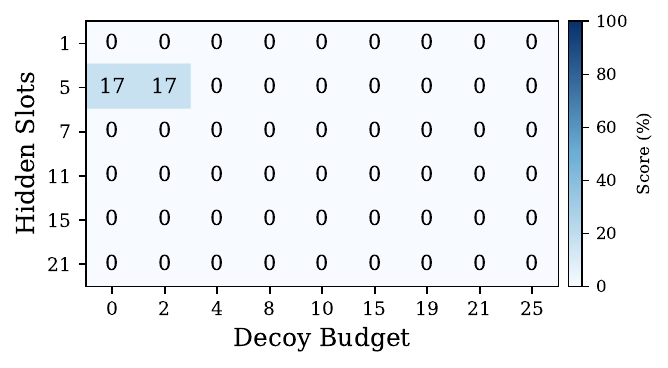}
        \caption{Qwen3.5-0.8B}
        \label{fig:heat_qwen_0p8b}
    \end{subfigure}
    \hfill
    \begin{subfigure}{0.32\linewidth}
        \centering
        \includegraphics[width=\linewidth]{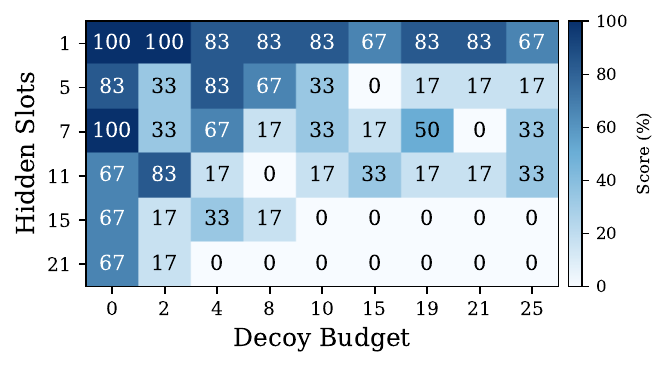}
        \caption{Qwen3.5-4B}
        \label{fig:heat_qwen_4b}
    \end{subfigure}
    \hfill
    \begin{subfigure}{0.32\linewidth}
        \centering
        \includegraphics[width=\linewidth]{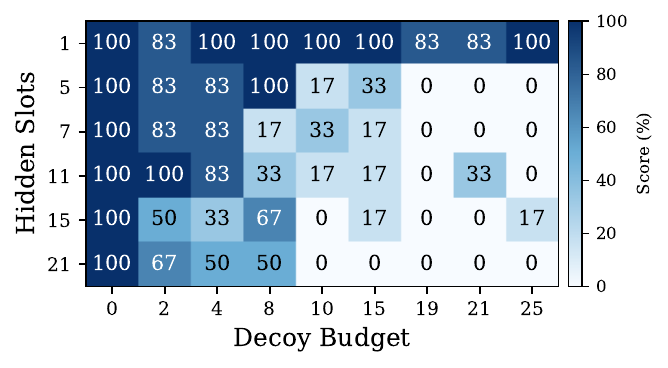}
        \caption{GLM-4.7-FP8}
        \label{fig:heat_glm}
    \end{subfigure}

    \vspace{0.5em}

    \begin{subfigure}{0.32\linewidth}
        \centering
        \includegraphics[width=\linewidth]{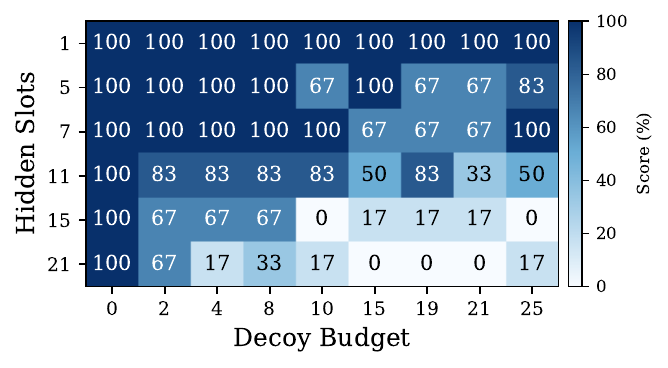}
        \caption{Qwen3.5-35B-A3B}
        \label{fig:heat_qwen_35b}
    \end{subfigure}
    \hfill
    \begin{subfigure}{0.32\linewidth}
        \centering
        \includegraphics[width=\linewidth]{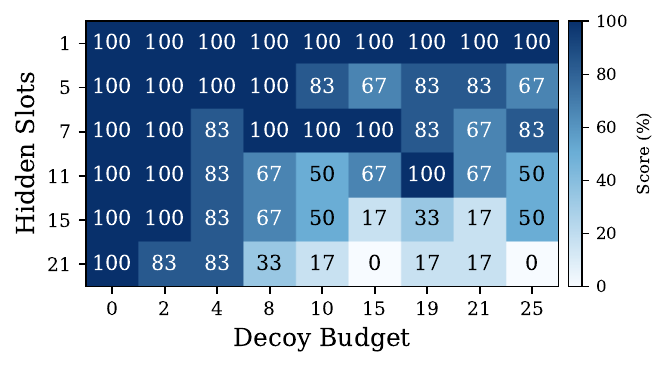}
        \caption{Qwen3.5-122B-A10B}
        \label{fig:heat_qwen_122b}
    \end{subfigure}
    \hfill
    \begin{subfigure}{0.32\linewidth}
        \centering
        \includegraphics[width=\linewidth]{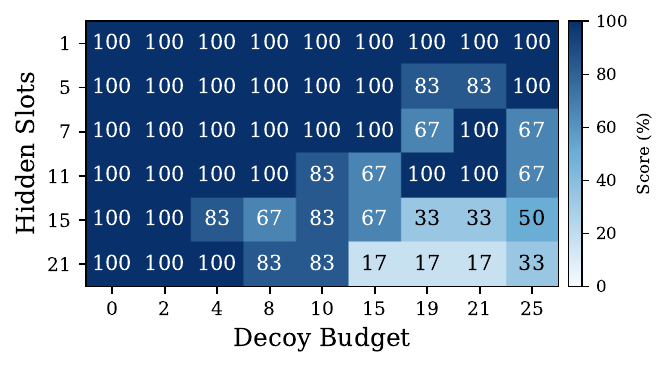}
        \caption{Qwen3.5-397B-A17B}
        \label{fig:heat_qwen_397b}
    \end{subfigure}

    \vspace{0.5em}

    \begin{subfigure}{0.32\linewidth}
        \centering
        \includegraphics[width=\linewidth]{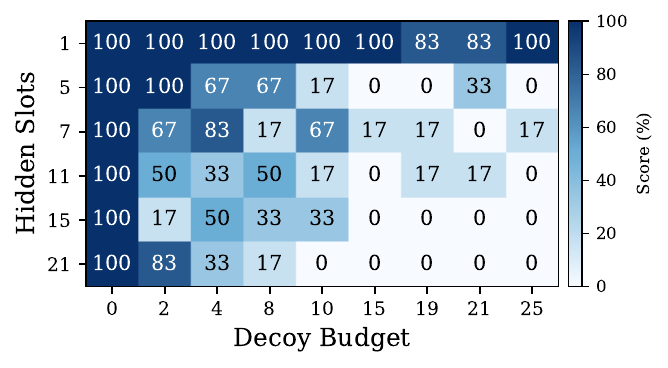}
        \caption{MiniMax-M2}
        \label{fig:heat_minimax_m2}
    \end{subfigure}
    \hfill
    \begin{subfigure}{0.32\linewidth}
        \centering
        \includegraphics[width=\linewidth]{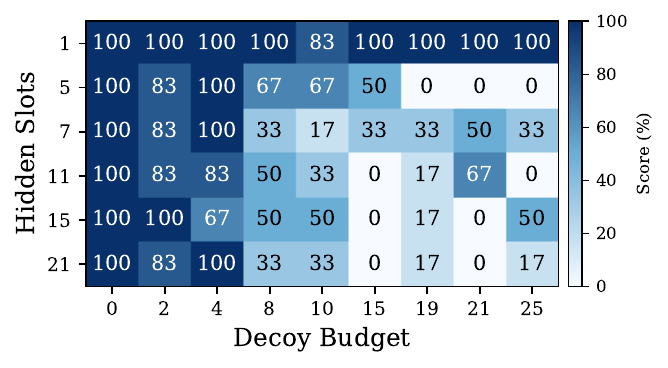}
        \caption{MiniMax-M2.1}
        \label{fig:heat_minimax_m2p1}
    \end{subfigure}
    \hfill
    \begin{subfigure}{0.32\linewidth}
        \centering
        \includegraphics[width=\linewidth]{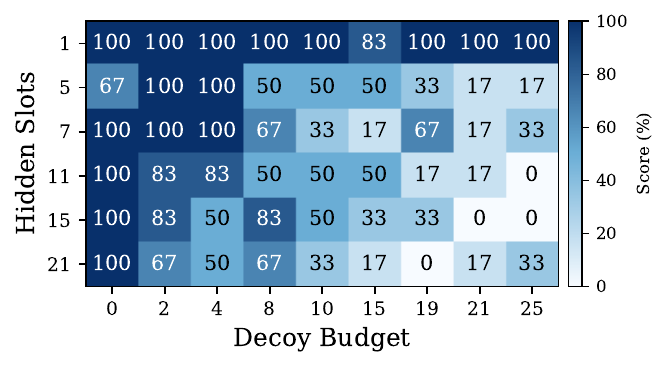}
        \caption{MiniMax-M2.5}
        \label{fig:heat_minimax_m2p5}
    \end{subfigure}

    \vspace{-8pt}
    \caption{Heatmaps of average reward (\%) across hidden slots and decoy budgets for all evaluated models. Higher reward indicates better task completion under the corresponding horizon and difficulty setting.}
    \label{fig:heatmaps_all}
\end{figure}

\subsubsection{Experimental Setup}

\textbf{Dataset Setup.}
The benchmark consists of six domains: \textit{course}, \textit{meal}, \textit{pc\_build}, \textit{shopping}, \textit{travel}, and \textit{workforce}. All domains share the same configuration: each domain contains 54 instances, each instance is a $5 \times 7$ grid (5 rows $\times$ 7 columns, yielding 35 slots per instance). Hidden slots are sampled from $h \in \{1, 5, 7, 11, 15, 21\}$ and decoy budgets from $b \in \{0, 2, 4, 8, 10, 15, 19, 21, 25\}$, resulting in $6 \times 9 = 54$ combinations of $(h, b)$ per domain. Each hidden slot has 25 candidates, among which one is the truth, $b$ are decoy candidates, and the remaining are filtered candidates. In total, the benchmark comprises $54 \times 6 = 324$ instances across all domains.

\textbf{Model Setup.}
We evaluate a diverse set of open-source models spanning different architectures and scales. For the \textit{Qwen3.5 Dense} family, we include five sizes: 0.8B, 2B, 4B, 9B, and 27B. For the \textit{Qwen3.5 MoE} family, we evaluate three variants: 35B-A3B, 122B-A10B, and 397B-A17B. We also include three models from the \textit{MiniMax} family (M2-229B, M2.1-229B, M2.5-229B), all based on MoE architecture. Additionally, we evaluate \textit{MiroThinker-mini}\citep{miromind2025mirothinker} (30B, Dense) and \textit{GLM-4.7-FP8}\citep{zeng2026glm} (358B, MoE). In total, 13 models are evaluated, covering a wide range of model sizes (0.8B to 397B active parameters) and architectures (Dense and MoE), enabling a comprehensive assessment of agent capability on the benchmark.

\textbf{Resource Setup.}
All experiments are conducted on a server equipped with 8$\times$H200 GPUs with 64 parallel threads. Each run uses a fixed random seed of 42 for reproducibility. The maximum number of interaction steps per instance is set to 600. Each model response is limited to a maximum of 16,384 output tokens; if this limit is exceeded more than three times within a single instance, the run is marked as failed.

\subsubsection{Results and Analysis}

\textbf{Overall Performance.}
\Cref{tab:main_results} presents the evaluation results across all models and domains. Among the evaluated models, Qwen3.5-397B-A17B achieves the highest overall score of 84.9\%. Within the Qwen3.5 Dense family, performance scales consistently with model size, from 0.6\% at 0.8B to 74.7\% at 27B. The MiniMax family achieves moderate performance (44.1\%--58.6\%), while MiroThinker-mini and Qwen3.5-0.8B/2B struggle significantly, suggesting that smaller or less capable models find the benchmark highly challenging.

\textbf{Heatmap Analysis.}
\Cref{fig:heatmaps_all} shows per-model heatmaps across hidden slots and decoy budgets. Weaker models (e.g., Qwen3.5-0.8B) score near zero across almost all settings, while stronger models degrade as $b$ or $h$ increases, confirming that the benchmark effectively differentiates model capabilities across both horizon and difficulty dimensions.

\subsection{Ablation: Tool Failure Simulation}

\begin{figure}[t]
    \centering
   \begin{subfigure}{0.32\linewidth}
        \centering
        \includegraphics[width=\linewidth]{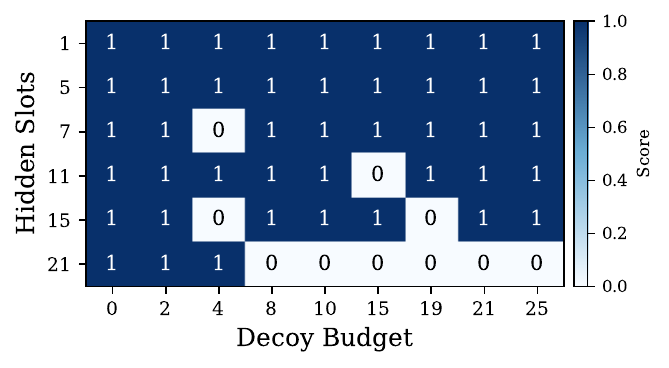}
        \caption{Fail rate = 0.0}
        \label{fig:failrate0.0}
    \end{subfigure}
    \hfill
    \begin{subfigure}{0.32\linewidth}
        \centering
        \includegraphics[width=\linewidth]{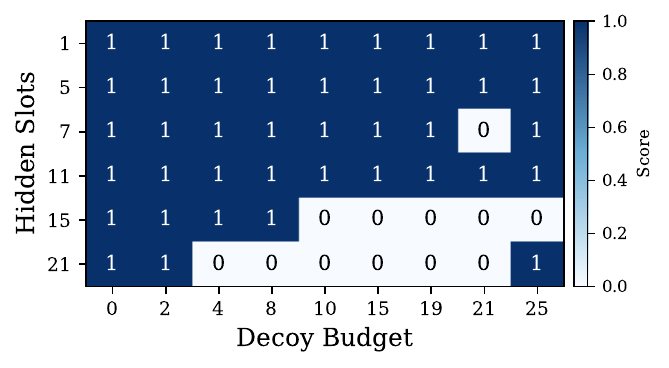}
        \caption{Fail rate = 0.1}
        \label{fig:failrate0.1}
    \end{subfigure}
    \hfill
    \begin{subfigure}{0.32\linewidth}
        \centering
        \includegraphics[width=\linewidth]{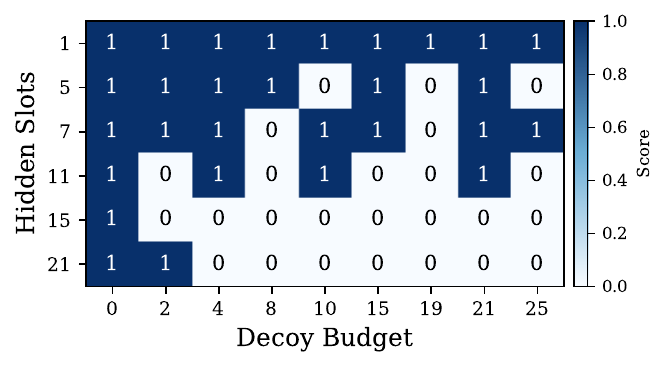}
        \caption{Fail rate = 0.3}
        \label{fig:failrate0.3}
    \end{subfigure}
    \vspace{-8pt}
    \caption{Heatmaps of average reward across hidden slots and decoy budgets on the Course domain under varying tool failure rates (0.0, 0.1, 0.3) of Qwen3.5-122B-A10B, simulating real-world tool call failures due to network or service instability. Higher failure rates consistently degrade agent performance across all horizon and difficulty levels.}
    \label{fig:tool_failure}
\end{figure}

To evaluate agent robustness under realistic conditions, we simulate tool call failures by randomly rejecting tool calls with probability $p \in \{0.0, 0.1, 0.3\}$, mimicking real-world instability such as network timeouts or service errors.

As shown in \Cref{fig:tool_failure}, even a modest failure rate of $p{=}0.1$ noticeably degrades performance across all horizon and difficulty levels, and the effect becomes more pronounced at $p{=}0.3$. Tasks with larger hidden slots and higher decoy budgets are particularly sensitive to tool failures. These results confirm that our benchmark can serve as a testbed for evaluating agent resilience beyond task-solving accuracy alone.



\section{Related Works}
\textbf{Agent Benchmark}\citep{yehudai2025survey}. A broad range of benchmarks has been proposed to evaluate LLM-based agents across different application domains. Web agent benchmarks\citep{zhou2023webarena,boisvert2024workarena++,drouin2024workarena,koh2024visualwebarena} focus on browser-based task completion, ranging from early environments. Software engineering benchmarks assess code generation and repository-level problem solving, with SWE-bench\citep{jimenez2023swe} and its variants\citep{deng2025swe,prathifkumar2025does,aleithan2024swe}. Scientific agent benchmarks such as ScienceWorld\citep{wang2022scienceworld,nathani2025mlgym,jansen2024discoveryworld,laurent2024lab} target research-oriented reasoning and experimentation tasks. Conversational agent benchmarks including $\tau$-Bench\citep{yao2024tau,barres2025tau,levi2025intellagent,castillo2024beyond} evaluate agents in dialogue-driven, tool-assisted service scenarios.

\textbf{Scalable Long-context Evaluation.} Early benchmarks such as LongBench\citep{bai2024longbench,bai2025longbench} and L-Eval\citep{an2024eval} established the foundation for evaluating long-context language model capabilities. Subsequent work such as $\infty$-Bench\citep{zhang2024infty} further extended context lengths to push model limits. Synthetic task-based benchmarks then emerged---most notably NIAH (Needle In A Haystack)\citep{yu2025sequential} and Ruler\citep{hsieh2024ruler}. More recently, benchmarks\citep{yang2025100} such as HELMET\citep{yen2024helmet} and LV-Eval\citep{yuan2024lv} introduced controllable context lengths alongside LLM-based evaluation metrics, enabling more systematic and fine-grained analysis.

\section{Conclusion}

We present \AgentCE, a lightweight agent benchmark with fine-grained control over task horizon and difficulty via hidden slots $H$ and decoy budget $B$. By resolving all tool calls through static JSON files, \AgentCE eliminates environment overhead while remaining fast, reproducible, and suitable for training-time validation. Comprehensive experiments across 13 models and 6 domains confirm that performance degrades consistently with increasing $H$ and $B$, demonstrating that \AgentCE provides interpretable and controllable evaluation of agent reasoning capabilities.



\newpage


\bibliography{colm2026_conference}
\bibliographystyle{colm2026_conference}

\clearpage

\appendix
\section{Dataset Generation Pipeline}
\label{appendix:data_generation}

This section provides a detailed description of the dataset generation procedure summarized in \Cref{alg:dataset_generation}.
The pipeline proceeds in two stages: (1) constructing a ground-truth solution and its associated constraints, and (2) generating candidate sets for each hidden slot.

\begin{algorithm}[ht]
\caption{Dataset Generation for Domain $d$}
\label{alg:dataset_generation}
\begin{algorithmic}[1]
\Require grid size $(R, C)$, hidden slots $H$, decoy budget $B$, candidates per slot $K$
\Ensure Instance with truth solution, constraints, and candidates
\State Sample truth grid $\mathcal{T} \in \mathcal{I}^{R \times C}$; generate global constraints $\mathcal{G}$
\State Select $H$ hidden slots $\mathcal{S}_H$; generate slot constraints $\mathcal{C}_s$ for each $s \in \mathcal{S}_H$
\State Select decoy slots $\mathcal{S}_B \subseteq \mathcal{S}_H$; partition $B$ into $\{b_s\}_{s \in \mathcal{S}_B}$
\For{each slot $s \in \mathcal{S}_H$ in decoy order}
    \State Collect filter candidates $\mathcal{F}_s$: items violating $\mathcal{C}_s$
    \If{$s \in \mathcal{S}_B$}
        \For{each attempt up to \texttt{max\_retries}}
            \State Sample $c$ satisfying $\mathcal{C}_s$ via targeted sampling on $\mathcal{G}$
            \If{$\forall \mathcal{H} \in \mathcal{H}_{\text{prior}}$: $\mathcal{H} \cup \{c\} \cup \mathcal{T}_{\text{future}} \Rightarrow \mathcal{G}$ violated}
                \Comment{hard constraint; $\mathcal{H}_{\text{prior}}$: all truth/decoy combos of prior decoy slots}
                \If{$s$ is not the last decoy slot \textbf{and} open-prefix preference holds}
                    \Comment{soft preference: $\mathcal{H} \cup \{c\} \cup \{\texttt{None}\}_{\text{future}}$ satisfies partial $\mathcal{G}$; relaxed over 3 levels}
                    \State Add $c$ to $\mathcal{D}_s$; \textbf{break}
                \ElsIf{$s$ is the last decoy slot}
                    \State Add $c$ to $\mathcal{D}_s$; \textbf{break}
                \Else
                    \State Relax soft preference (3 levels); retry
                \EndIf
            \EndIf
        \EndFor
    \EndIf
    \State $\text{candidates}_s \leftarrow \{\mathcal{T}[s]\} \cup \mathcal{D}_s \cup \mathcal{F}_s$, padded to $K$
\EndFor
\State \Return partial solution, $\mathcal{G}$, $\{\mathcal{C}_s\}$, $\{\text{candidates}_s\}$, $\{\mathcal{D}_s\}$, $\{\mathcal{F}_s\}$
\end{algorithmic}
\end{algorithm}

\subsection{Stage 1: Truth Solution and Constraints}

We begin by sampling a complete truth grid $\mathcal{T} \in \mathcal{I}^{R \times C}$ that serves as the ground-truth answer.
Global constraints $\mathcal{G}$ are then derived from $\mathcal{T}$, encoding conditions that the entire grid must satisfy (e.g., upper bounds on total credits, limits on category repetition).
Slot-level constraints $\mathcal{C}_s$ are generated \emph{only} for hidden slots $s \in \mathcal{S}_H$, where each $\mathcal{C}_s$ typically covers two item attributes, and the union of all slot constraints is designed to span the full attribute space.
A subset $\mathcal{S}_B \subseteq \mathcal{S}_H$ is designated as \emph{decoy slots}, and the decoy budget $B$ is partitioned into per-slot allocations $\{b_s\}_{s \in \mathcal{S}_B}$.

\subsection{Stage 2: Candidate Generation}

For each hidden slot $s$, the candidate set $\text{candidates}_s$ is composed of three disjoint parts:

\begin{itemize}
    \item \textbf{Truth candidate} $\mathcal{T}[s]$: the ground-truth item for slot $s$.
    \item \textbf{Filter candidates} $\mathcal{F}_s$: items that \emph{directly violate} the local slot constraint $\mathcal{C}_s$. These serve as obvious distractors that a capable agent should eliminate through local reasoning alone.
    \item \textbf{Decoy candidates} $\mathcal{D}_s$: items that \emph{satisfy} $\mathcal{C}_s$ locally but are designed to trigger a global constraint violation when selected. Decoys require multi-step global reasoning to identify and are the primary source of benchmark difficulty.
\end{itemize}

\noindent The final candidate set is padded to a fixed size $K$ and presented to the agent alongside the partial solution and constraints.

\subsection{Decoy Generation}

Generating valid decoy candidates is the most technically involved step of the pipeline.
Each decoy $c$ for a decoy slot $s$ must satisfy two conditions.

\subsubsection*{Hard Constraint (Global Invalidity Guarantee)}

The candidate $c$ must cause a global constraint violation regardless of the decisions made at \emph{prior} decoy slots.
Formally, for every combination $\mathcal{H}$ of previously generated truth or decoy assignments to earlier decoy slots, selecting $c$ at slot $s$ while filling all future hidden slots with their truth values must violate $\mathcal{G}$:
\[
    \forall \mathcal{H} \in \mathcal{H}_{\text{prior}}: \quad \mathcal{H} \cup \{c\} \cup \mathcal{T}_{\text{future}} \Rightarrow \mathcal{G} \text{ violated.}
\]
This ensures that selecting $c$ \emph{at any point along any prior decision path} will ultimately lead to a globally invalid solution.

\subsubsection*{Soft Preference (Open-Prefix Validity)}

Beyond the hard constraint, decoys should ideally remain \emph{locally undetectable} when future hidden slots are not yet filled.
We define open-prefix validity as the condition that the current partial assignment (history $\cup$ $\{c\}$ $\cup$ future $= \texttt{None}$) does not prematurely expose a global violation.
To balance decoy quality against sampling tractability, the generator applies a three-level preference relaxation scheme controlled by thresholds $\langle t_1, t_2, t_3 \rangle$ (default: $\langle 30, 50, 70 \rangle$ retries):

\begin{itemize}
    \item \textbf{Level 1} (attempts $\leq t_1$): require open-prefix validity under \emph{all} historical truth/decoy combinations.
    \item \textbf{Level 2} (attempts $\leq t_2$): relax to a prefix-truth / suffix-decoy historical pattern.
    \item \textbf{Level 3} (attempts $\leq t_3$): only require open-prefix validity when all prior decoy slots remain at their truth values.
    \item \textbf{Beyond $t_3$}: drop the soft preference entirely; only the hard constraint is enforced.
\end{itemize}

\noindent Intuitively, a decoy found earlier (under stricter preferences) is more deceptive, as it remains superficially valid across a wider range of partial states.
Note that when checking global validity with future slots set to $\texttt{None}$, only \emph{upper-bound} type constraints are enforced, since lower-bound constraints (e.g., ``at least $k$ credits'') may still be satisfied once future slots are filled with truth values.

\subsection{Targeted Sampling}

To avoid pure rejection sampling, the generator uses \emph{targeted sampling}: it pre-computes target specifications from the current historical context and $\mathcal{G}$, prioritizing upper-bound constraints (e.g., total credit caps, category repetition limits) as the basis for decoy construction.
Such constraints naturally yield candidates that are globally invalid when future truth values are filled in (causing the bound to be exceeded) while remaining temporarily valid when future slots are empty (as the running total is lower).
For the final decoy slot, the three-level open-prefix preference is skipped entirely and only the hard constraint is enforced, avoiding degenerate sampling scenarios where the open-prefix condition is unsatisfiable by construction.

\subsection{Validation}

Each generated instance is automatically validated to ensure: (i) the truth solution satisfies all slot and global constraints; (ii) each hidden slot has the correct number of candidates; and (iii) all decoy candidates satisfy the multi-stage global invalidity guarantees described above.
Instances that fail validation are resampled; if the maximum retry count is exceeded, an exception is raised and the failure condition is reported.

\section{Tool System Design}
\label{appendix:tool_system}

The benchmark provides agents with a structured set of tools to interact with the grid-based constraint satisfaction tasks.
The tool system follows a two-tier design: a set of \emph{domain-agnostic common tools} shared across all domains, and a set of \emph{domain-specific tools} that expose the semantics of each individual domain.
This separation keeps the interaction interface clean and uniform while allowing the system to scale to new domains with minimal overhead.

\subsection{Common Tools}

Six tools are available in all domains regardless of the task context.
These tools operate solely on the grid structure and budget state, and carry no assumptions about the type of items being placed.
Table~\ref{tab:common_tools} summarizes their functionality.

\begin{table}[ht]
\centering
\caption{Common tools available across all domains.}
\label{tab:common_tools}
\small
\begin{tabular}{lp{7cm}}
\toprule
\textbf{Tool} & \textbf{Description} \\
\midrule
\texttt{set\_slot(row, col, id)} & Place an item \texttt{id} into the grid at \texttt{(row, col)}, or clear the slot if \texttt{id} is \texttt{None}. \\
\texttt{get\_current\_grid\_state()} & Return the full current state of the grid, including all filled and empty slots. \\
\texttt{get\_slot\_id(row, col)} & Query the item \texttt{id} currently occupying a specific grid cell. \\
\texttt{get\_hidden\_slot\_query\_budget(row, col)} & Return the remaining attribute query budget for a given hidden slot. \\
\texttt{get\_global\_check\_budget()} & Return the remaining budget for global constraint checks. \\
\texttt{done()} & Signal that the agent has completed the task. \\
\bottomrule
\end{tabular}
\end{table}

The common tools cover three orthogonal concerns: \emph{grid manipulation} (\texttt{set\_slot}), \emph{state inspection} (\texttt{get\_current\_grid\_state}, \texttt{get\_slot\_id}), and \emph{budget tracking} (\texttt{get\_hidden\_slot\_query\_budget}, \texttt{get\_global\_check\_budget}, \texttt{done}).
Because none of these operations depend on item semantics, they are naturally shared across all domains.

\subsection{Domain-Specific Tools}

Each domain provides five additional tools that expose item-level information and constraint checking for that domain.
Following a consistent naming convention \texttt{\{action\}\_\{domain\}\_\{function\}}, these tools form a uniform interface regardless of the underlying domain.
Table~\ref{tab:domain_tools} describes the five tool types.

\begin{table}[ht]
\centering
\caption{Domain-specific tools (instantiated per domain by replacing \texttt{\{domain\}} with the domain name).}
\label{tab:domain_tools}
\small
\begin{tabular}{p{6.5cm}p{7cm}}
\toprule
\textbf{Tool} & \textbf{Description} \\
\midrule
\texttt{query\_\{domain\}\_candidate\_from\_attribute(row, col, field, operator, value)}
& Filter the candidate list for a given slot by an attribute condition \texttt{(field~operator~value)}, narrowing the search space. \\
\addlinespace
\texttt{get\_\{domain\}\_item\_info(id)}
& Retrieve the full attribute profile of a single item. \\
\addlinespace
\texttt{get\_\{domain\}\_item\_attributes(ids, field)}
& Batch-query a specific attribute field for a list of item ids. \\
\addlinespace
\texttt{check\_\{domain\}\_slot\_constraints(row, col)}
& Check whether the item currently placed at \texttt{(row, col)} satisfies the local slot constraints for that position. \\
\addlinespace
\texttt{check\_\{domain\}\_global\_constraints()}
& Check whether the entire current grid satisfies all global constraints. \\
\bottomrule
\end{tabular}
\end{table}

The five tool types cover the full reasoning cycle an agent must perform: \emph{candidate filtering} (\texttt{query}), \emph{item inspection} (\texttt{get\_item\_info}, \texttt{get\_item\_attributes}), and \emph{constraint verification} (\texttt{check\_slot\_constraints}, \texttt{check\_global\_constraints}).

The benchmark currently supports six domains, each representing a distinct real-world scheduling or selection scenario:

\begin{itemize}
    \item \textbf{course}: an academic course selection task, where the agent fills a curriculum grid subject to credit limits and prerequisite-style constraints.
    \item \textbf{shopping}: a product selection task, where items must be chosen from a catalog to satisfy budget and category constraints.
    \item \textbf{travel}: an itinerary planning task, where the agent selects destinations or activities subject to time and cost constraints.
    \item \textbf{workforce}: a staff assignment task, where employees are allocated to roles under skill and workload constraints.
    \item \textbf{meal}: a meal planning task, where dishes are arranged into a weekly plan satisfying nutritional and dietary constraints.
    \item \textbf{pc\_build}: a computer assembly task, where hardware components must be selected to meet compatibility and budget constraints.
\end{itemize}

\subsection{Design Rationale}

The two-tier architecture provides two key advantages.
First, \emph{interface uniformity}: agents interact with all domains through the same set of tool signatures, meaning a general-purpose agent policy requires no domain-specific adaptation to operate across tasks.
Second, \emph{extensibility}: adding a new domain requires only implementing the five domain-specific tools; the common tools and the overall evaluation framework remain unchanged.
Together, these properties make the benchmark straightforward to scale and easy to integrate with diverse agent architectures.

\end{document}